\documentclass{article}

\usepackage[final, nonatbib]{nips_2017}
\usepackage{subcaption}
\usepackage{varwidth}
\usepackage[utf8]{inputenc} 
\usepackage[T1]{fontenc}    
\usepackage{hyperref}       
\usepackage{url}            
\usepackage{booktabs}       
\usepackage{amsfonts}       
\usepackage{nicefrac}       
\usepackage{microtype}      
\usepackage{graphicx}
\usepackage{algorithm}
\usepackage{algorithmic}

\usepackage{amsfonts}
\usepackage{amsmath}
\newsavebox\tmpbox

\newcommand{\remove}[1]{}

\title{Generative Compression}

\author{
  Shibani Santurkar \quad David Budden$^\dagger$ \quad Nir Shavit  \\
  Massachusetts Institute of Technology\\
  \texttt{\{shibani,budden,shanir\}@mit.edu} \\
}

\begin{document}

\maketitle

\begin{abstract}
  Traditional image and video compression algorithms rely on hand-crafted encoder/decoder pairs (codecs) that lack adaptability and are agnostic to the data being compressed. Here we describe the concept of {\it generative compression}, the compression of data using generative models, and suggest that it is a direction worth pursuing to produce more accurate and visually pleasing reconstructions at much deeper compression levels for both image and video data. We also demonstrate that generative compression is orders-of-magnitude more resilient to bit error rates (e.g. from noisy wireless channels) than traditional variable-length coding schemes.
\end{abstract}
\section{Introduction}
\label{sec:intro}
\begin{figure}[h!]
	\begin{center}
		\centerline{\includegraphics[width=0.95\columnwidth]{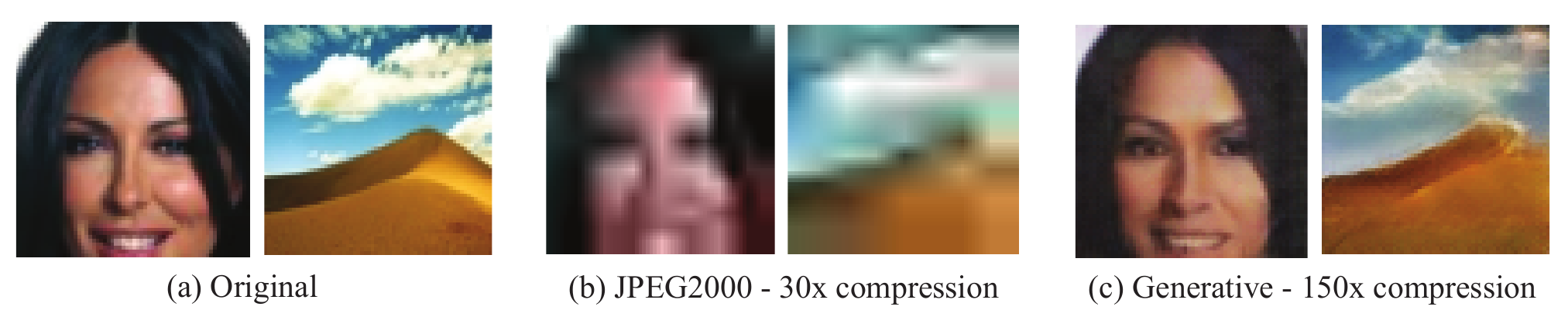}}
		\caption{Traditional versus generative image compression.}
		\label{fig:introfig}
	\end{center}
	\vskip -0.2in
\end{figure}

Graceful degradation is a quality-of-service term used to capture the idea that, as bandwidth drops or transmission errors occur, user experience deteriorates but continues to be meaningful. Traditional compression techniques, such as JPEG, are agnostic to the data being compressed and do not degrade gracefully. This is shown in Figure~\ref{fig:introfig}, which compares (a) two original images to (b) their JPEG2000-compressed representations. Building upon the ideas of~\cite{gregor2016towards} and the recent promise of deep generative models~\cite{goodfellow2014generative}, this paper presents a framework for \emph{generative compression} of image and video data. As seen in Figure~\ref{fig:introfig}(c), this direction shows great potential for compressing data so as to provide graceful degradation, and to do so at bandwidths far beyond those reachable by traditional techniques.

There are two main categories of data compression, descriptively named \emph{lossless} and \emph{lossy}. The former problem traditionally involved deriving codes for discrete data given knowledge of their underlying distribution, the entropy of which imposes a bound on achievable compression. To deliver graceful degradation, we focus on the relaxed problem of lossy compression, where we believe there is potential for orders-of-magnitude improvement using generative compression compared to existing algorithms. Too see why, consider the string $s = \mathtt{grass\,tennis\,court}$. This string contains just a few bytes of information, and yet the detail and vividity of your mental reconstruction is astounding. Likewise, an MNIST-style $28$x$28$ grayscale image can represent many more unique images than there are atoms in the universe. How small of a region of this space is spanned by plausible MNIST samples? The promise of generative compression is to translate this perceptual redundancy into a reduction in code verbosity.

Lossy compression has traditionally been formulated as a rate-distortion optimization problem. In this framework, an \emph{analysis} transform, $f : \mathbb{R}^N \to \mathbb{R}^M$, maps input data $\textbf{x}$ (e.g. a vector of pixel intensities) to a vector $\mathbf{z}$ in latent code space, and a \emph{synthesis} transform, $g : \mathbb{R}^M \to \mathbb{R}^N$, maps $\mathbf{z}$ back into the original space. Compression is achieved by (lossy) quantization of $\mathbf{z}$ followed by lossless compression using an entropy coding scheme. In this form, compression seeks to minimize both the rate of the latent code, lower-bounded by the entropy of its distribution, and the distortion of the output, typically reported as a signal-to-noise-ratio (PSNR) or structural similarity (SSIM) metric.

Joint optimization over rate and distortion has long been considered an intractable problem for images and other high-dimensional spaces~\cite{gersho1992vector}. Attention has instead been focused on hand-crafting encoder/decoder pairs (codecs) that apply linear analysis and synthesis transforms, e.g. discrete cosine transforms (JPEG) and multi-scale orthogonal wavelet decomposition (JPEG2000). There are several limitations to this approach. There is no reason to expect that a linear function is optimal for compressing the full spectrum of natural images. Even presuming they are optimal for a particular class of bitmap images, this performance is unlikely to generalize to emerging media formats, and the development and standardization of new codecs has historically taken many years. 
\section{Generative Models for Image Compression}
\label{sec:generative}
\remove{ 
Here we make liberal use of standard terminology (e.g. ``manifold") to explain our compression methodology. Consider the top-down argument presented in Section 1, such that $X$ is the 784-dimensional space of $28$x$28$ grayscale images. Let us further assume that all MNIST samples of handwritten digits cluster together within some smaller subspace, $Z$. Presented with a new MNIST vector, $\mathbf{x}$, one approach to compression is to represent $\mathbf{x}$ by the lower-dimensional coordinate, $\mathbf{z}$, of its orthogonal projection onto manifold $Z$. Reconstructing the original image is the reverse process of mapping $\mathbf{z}$ to $\hat{\mathbf{x}} \in X$. The intuition behind this framework is that the distortion loss, $L(\mathbf{x}, \hat{\mathbf{x}})$, measures how poorly $\mathbf{x}$ is captured by our understanding of the latent structure of the manifold $Z$ of MNIST images.
} 
A pleasing alternative is to replace hand-crafted linear transforms with artificial neural networks, i.e. replacing the analysis transform with a learnt encoder function, $\textbf{z} = f_ \theta(\textbf{x})$, and the synthesis transform with a learnt decoder function, $\hat{\textbf{x}} = g_\phi(\textbf{z})$. Noteworthy examples include the compressive autoencoder~\cite{Theis2017a}, which derives differentiable approximations for quantization and entropy rate estimation to allow end-to-end training by gradient backpropagation. The authors of \cite{balle2016end} achieve a similar result, using a joint nonlinearity as a form of gain control. Also noteworthy is the LSTM-based autoencoder framework presented in~\cite{toderici2015variable}, specifically designed for the common failure case of compressing small thumbnail images. This approach was later extended using fully-convolutional networks for the compression of full-resolution images~\cite{toderici2016full}. Collectively, these and similar models are showing promising results in both lossless~\cite{oord2016pixel,theis2015generative} and lossy data compression~\cite{gregor2016towards,toderici2016full}. 

Recent advancements in generative modelling also show promise for compression. Imagine that the role of the receiver is simply to synthesize some realistic looking MNIST sample. If we knew the true distribution, $P(\mathbf{x})$, of this class of images defined over $X$, we could simply sample $\hat{\mathbf{x}} \in \mathbb{R}^N$ from this distribution. Unfortunately, it is intractable to accurately estimate this density function for such a high-dimensional space. One remedy to this problem is to factorize $P(\mathbf{x})$ as the product of conditional distributions over pixels. This sequence modeling problem can be solved effectively using autoregressive models of recurrent neural networks, allowing the generation of high-quality images or in-filling of partial occlusions~\cite{oord2016pixel}. However, these models forego a latent representation and as such do not provide a mechanism for decoding an image from a specific code. 

To implement our decoder, we can instead apply a generator function, $\hat{\mathbf{x}} = g_\phi(\mathbf{z})$, to approximate $P(\mathbf{x})$ as the transformation of some prior latent distribution, $P(\mathbf{z})$. To generate realistic-looking samples, we wish to train $g$ to minimize the difference between its distribution, $P_\phi(\mathbf{x}) = \mathbb{E}_{\mathbf{z}\sim P(\mathbf{z})}\, [P_\phi(\mathbf{x}|\mathbf{z})]$, and the unknown true distribution, $P(\mathbf{x})$. A popular solution to this problem is to introduce an auxiliary discriminator network, $d_\vartheta(\mathbf{x})$, which learns to map $\mathbf{x}$ to the probability that it was sampled from $P(\mathbf{x})$ instead of $P_\phi(\mathbf{x})$~\cite{goodfellow2014generative}. This framework of generative adversarial networks (GANs) simultaneously learns $\phi$ and $\vartheta$ by training against the minimax objective:
\begin{equation*}
\mathbb{E}_{\mathbf{x}\sim P(\mathbf{x})}\, [\log d(\mathbf{x})] + \mathbb{E}_{\mathbf{z}\sim P(\mathbf{z})}\, [\log(1- d(g(\mathbf{z}))].
\end{equation*}
The authors showed that this objective is equivalent to minimizing the Jensen-Shannon divergence between $P_\phi(\mathbf{x})$ and $P(\mathbf{x})$ for ideal discriminators. Although GANs provide an appealing method for reconstructing quality images from their latent code, they lack the \emph{inference} (encoder) function $f : X \to Z$ necessary for image compression. Points can be mapped from $Z$ to $X$, but not vice versa.

An alternative to GANs for generative image modeling are variational autoencoders~\cite{kingma2013auto}. Similar to GANs, VAEs introduce an auxiliary network to facilitate training. Unlike GANs, this inference function is trained to learn an approximation, $Q_{\theta}(\textbf{z}|\textbf{x})$, of the true posterior, $P(\textbf{z}|\textbf{x})$, and thus can be used as an encoder for image compression. This is achieved by maximizing the log-likelihood of the data under the generative model in terms of a variational lower bound, $\mathcal{L}(\theta,\phi;\mathbf{x})$. Recent studies have demonstrated the potential of VAEs for compression by training $P_\theta(\mathbf{x}|\mathbf{z})$ and $Q_\phi(\mathbf{z} | \mathbf{x})$ as deep neural networks. The authors of~\cite{gregor2016towards} report that they do not build an actual compression algorithm, but present sample reconstructions with perceptual quality similar to JPEG2000. However, a well-established limitation of VAEs (and autoencoders more generally) is that maximizing a Gaussian likelihood is equivalent to minimizing the $L^2$ loss between pixel intensity vectors. This loss is known to correlate poorly with human perception and leads to blurry reconstructions~\cite{theis2015generative,larsen2015autoencoding}.

\section{Neural Codecs for Generative Compression}
\label{sec:autoencoders}
\begin{figure}[!t]
\begin{subfigure}[b]{0.45\textwidth}
\begin{center}
\centerline{\includegraphics[width=1.0\columnwidth]{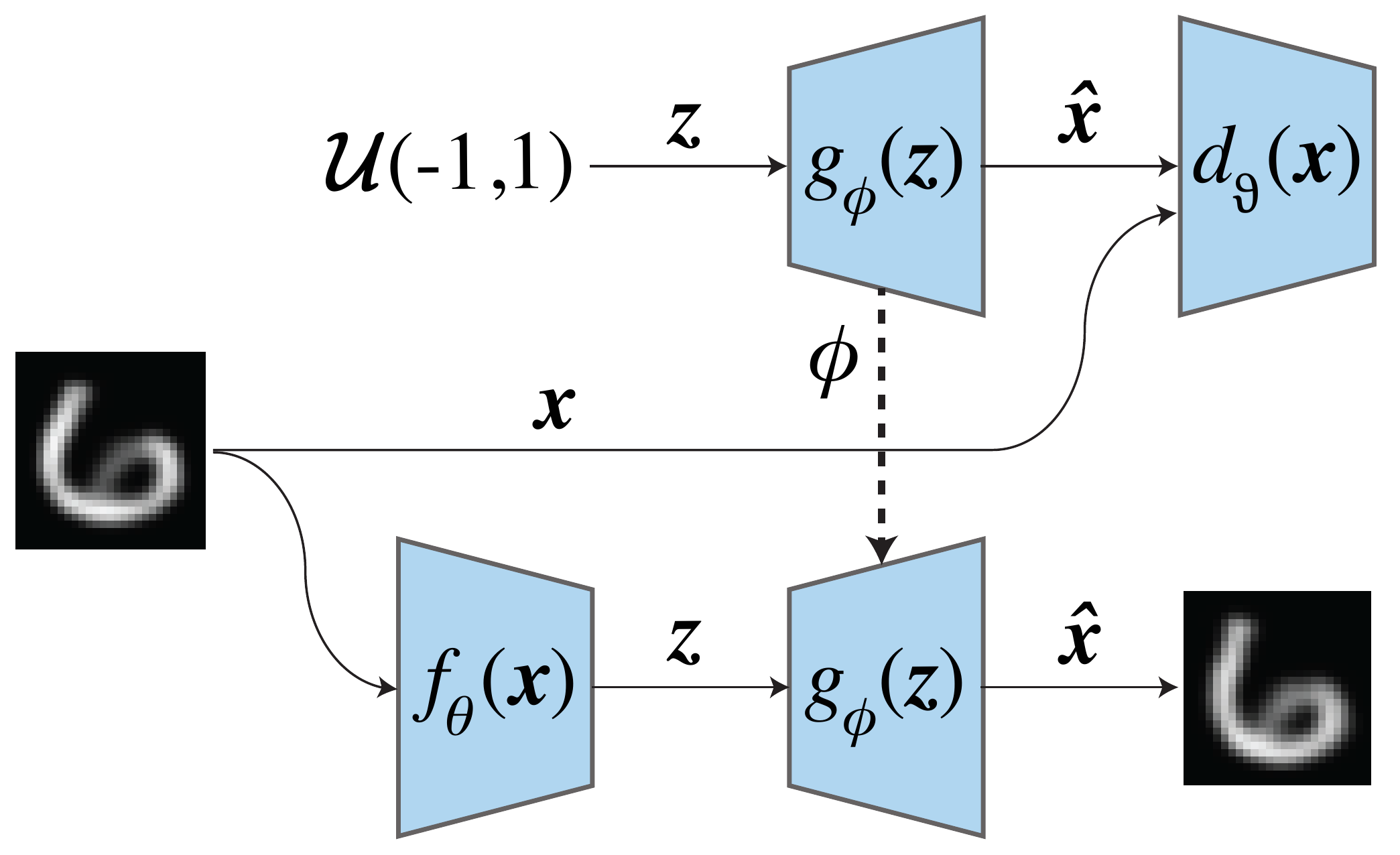}}
\end{center}
\end{subfigure}%
\hfill
\begin{subfigure}[b]{0.40\textwidth}
	\begin{center}
		\centerline{\includegraphics[width=1.0\columnwidth]{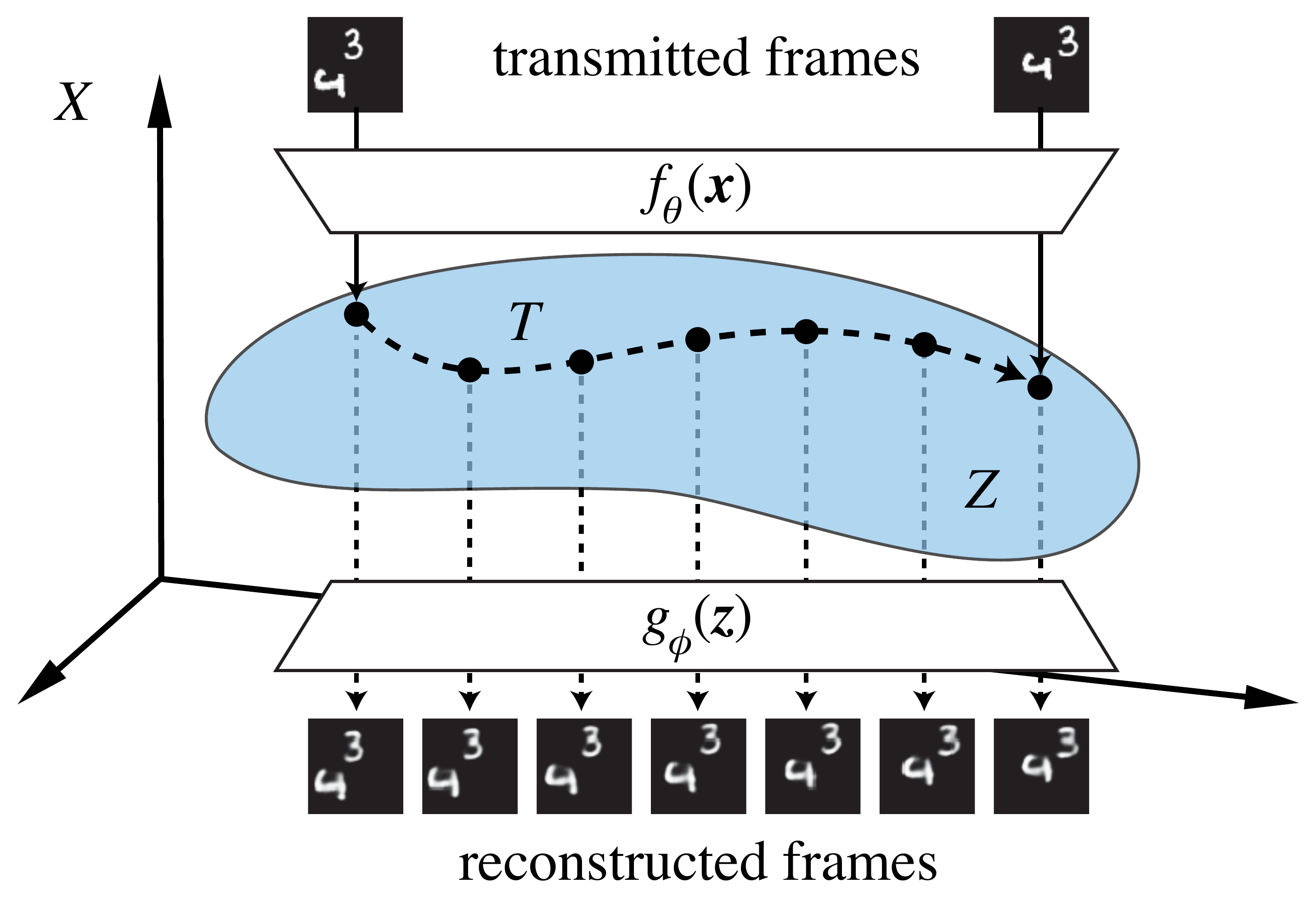}}
	\end{center}
\end{subfigure}%
\caption{Generative compression architectures for (left) image and (right) video data.}
\label{fig:sys}
\vskip -0.1in
\end{figure}

To build an effective neural codec for image compression, we implement the paired encoder/decoder interface of a VAE while generating the higher-quality images expected of a GAN. We propose a simple neural codec architecture (NCode,  Figure~\ref{fig:sys}a)  that approaches this in two stages. First, a decoder network, $g_\phi : Z \to X$, is greedily pre-trained using an adversarial loss with respect to the auxiliary discriminator network, $d_\vartheta : X \to [0,1]$. For this stage, $g_\phi$ and $d_\vartheta$ are implemented using DCGAN-style ConvNets~\cite{radford2015unsupervised}. Second, an encoder network, $f_\theta : X \to Z$, is trained to minimize some distortion loss, $L(\mathbf{x}, g_\phi(f_\theta(\mathbf{x})))$, with respect to this non-adaptive decoder. We also investigate methods for lossy quantization of $\mathbf{z}$, motivated by recent studies demonstrating the robustness of deep neural nets to reduced numerical precision~\cite{gupta2015deep,hubara2016quantized}. Compression is improved by either (a) reducing the length of the latent vector, and/or (b) reducing the number of bits used to encode each entry.

Traditional image compression algorithms have been crafted to minimize pixel-level loss metrics. Although optimizing for MSE can lead to good PSNR characteristics, the resulting images are perceptually implausible due to a depletion of high-frequency components (blurriness)~ \cite{ledig2016photo}. By adversarially pre-training a non-adaptive decoder, the codec will tend to produce samples more like those that fool a frequency-sensitive auxillary discriminator. To further improve the plausibility of our reconstructed images, we also choose to enrich the distortion loss with an additional measure of perceptual quality. Recent studies have indicated that textural information of an image is effectively captured by the feature maps of deep ConvNets pretrained for object recognition \cite{gatys2015texture}. Perceptual loss metrics derived from these features have been used to improve the plausibility of generative image models~\cite{dosovitskiy2016generating} and successfully applied to applications including super-resolution \cite{ledig2016photo} and style transfer \cite{gatys2016image}. We take a similar approach in NCode, modeling the distortion between image, $\mathbf{x}$, and reconstruction, $\mathbf{\hat{x}} = g_\phi(f_\theta(\mathbf{x}))$, as the weighted sum of pixel-level and perceptual losses:
\begin{equation*}
L(\mathbf{x}, \mathbf{\hat{x}}) = \lambda_1 || \mathbf{x} - \mathbf{\hat{x}} ||_2 \,+\, \lambda_2|| \text{conv}_4(\mathbf{x}) - \text{conv}_4(\mathbf{\hat{x}})||_2,
\end{equation*}
where $\text{conv}_4$ is the fourth convolutional layer of an ImageNet-pretrained AlexNet~\cite{krizhevsky2012imagenet, zhu2016generative}. 

Our NCode model architecture was largely motivated by the work of~\cite{zhu2016generative}, where a similar framework was applied for on-manifold photo editing. It also bares similarity to many other existing models. For example, recent studies have proposed hybrid models that combine VAEs with GANs~\cite{larsen2015autoencoding,dosovitskiy2016generating,lamb2016discriminative}. Our model differs in the adoption of a non-adaptive and adversarially pre-trained decoder, our hybrid perceptual/pixel loss function, and the use of a vanilla autoencoder architecture (in place of a VAE).  Similarly, other studies have augmented GANs with inference functions, e.g. adversarial feature learning~\cite{donahue2016adversarial} and adversarially learned inference (ALI)~\cite{dumoulin2016adversarially}. The ALI study also describes a similar compression pipeline, passing images through a paired encoder/decoder and assessing the reconstruction quality. Although the ALI generator produces high-quality and plausible samples, they differ dramatically in appearance from the input. Unlike our NCode model, the authors attribute this to a lack of explicit pixel-level distortion in their optimization objective.


\subsection{Generative Video Compression}
\label{ss:videocomp}
Here we present what is, to our knowledge, the first example of neural network-based compression of video data at sub-MPEG rates. As a video is simply a sequence of images, these images can be compressed and transmitted frame-by-frame using NCode. This is reminiscent of the motion-JPEG scheme in traditional video compression literature. However, this approach fails to capture the rich temporal correlations in natural video data~\cite{budden2016deep,shi2016real}. Instead, we would prefer our model to be inspired by the interpolation (bidirectional prediction) scheme introduced for the popular MPEG standard.

The simplest method of capturing temporal redundancy is to transmit only every $N$-th frame, $\mathbf{X} = [\mathbf{x}^{(t)},\, \mathbf{x}^{(t+N)},\, \mathbf{x}^{(t+2N)}, \dots]$, requiring the receiver to interpolate the missing data with a small $N$-frame latency. The traditional limitation of this approach is that interpolation in pixel-space yields visually displeasing results. Instead, we choose to model a video sequence as uniformly-spaced samples along a path, $T$, on the manifold, $Z$ (Figure~\ref{fig:sys}). We assume that $Z$ is a lower-dimensional embedding of some latent image class, and further that for sufficiently small $N$, the path $\mathbf{x}^{(t)} \rightarrow\mathbf{x}^{(t+N)}$ can be approximated by linear interpolation on $Z$. This assumption builds on the wealth of recent literature demonstrating that interpolating on manifolds learnt by generative models produce perceptually cohesive samples, even between quite dissimilar endpoints~\cite{radford2015unsupervised,zhu2016generative,brock2016neural}. 

{Similar to MPEG, we can further compress a video sequence through delta and entropy coding schemes (specifically, Huffman coding). Each latent vector is transmitted as its difference with respect to the previous transmitted frame, $\boldsymbol{\delta}^{(t+N)} = \mathbf{z}^{(t+N)} - \mathbf{z}^{(t)}$. We observe that this representation gains far more from entropy coding than for individual latent vectors sampled from $P(\mathbf{z})$, leading to a further (lossless) reduction in bitrate. We do not use entropy coding for NCode image compression.}


\section{Experiments}
\label{sec:experiments}
\begin{figure*}[!t] 
	\includegraphics[width=0.94\textwidth]{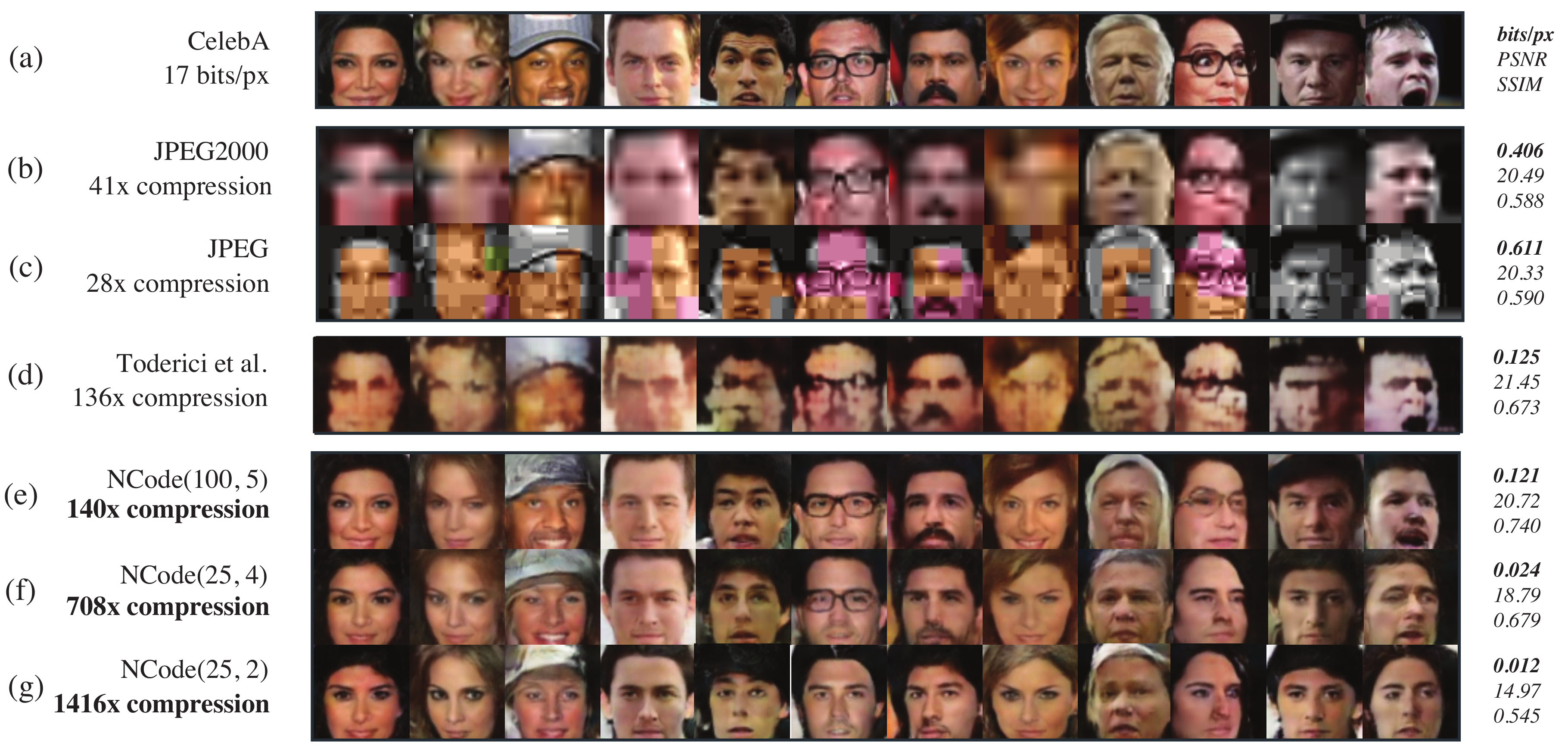}
	\includegraphics[width=0.94\textwidth]{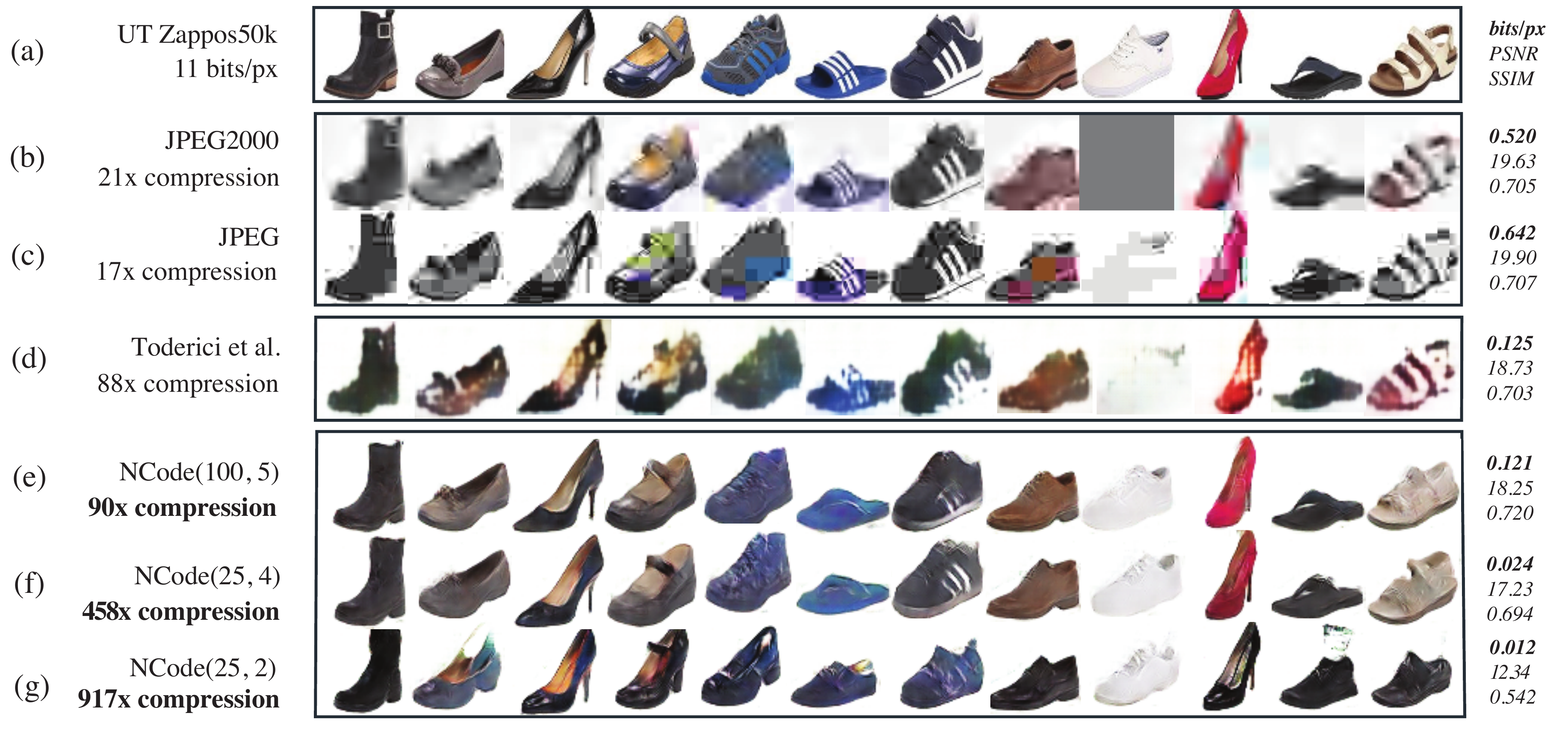}
	\includegraphics[width=0.94\textwidth]{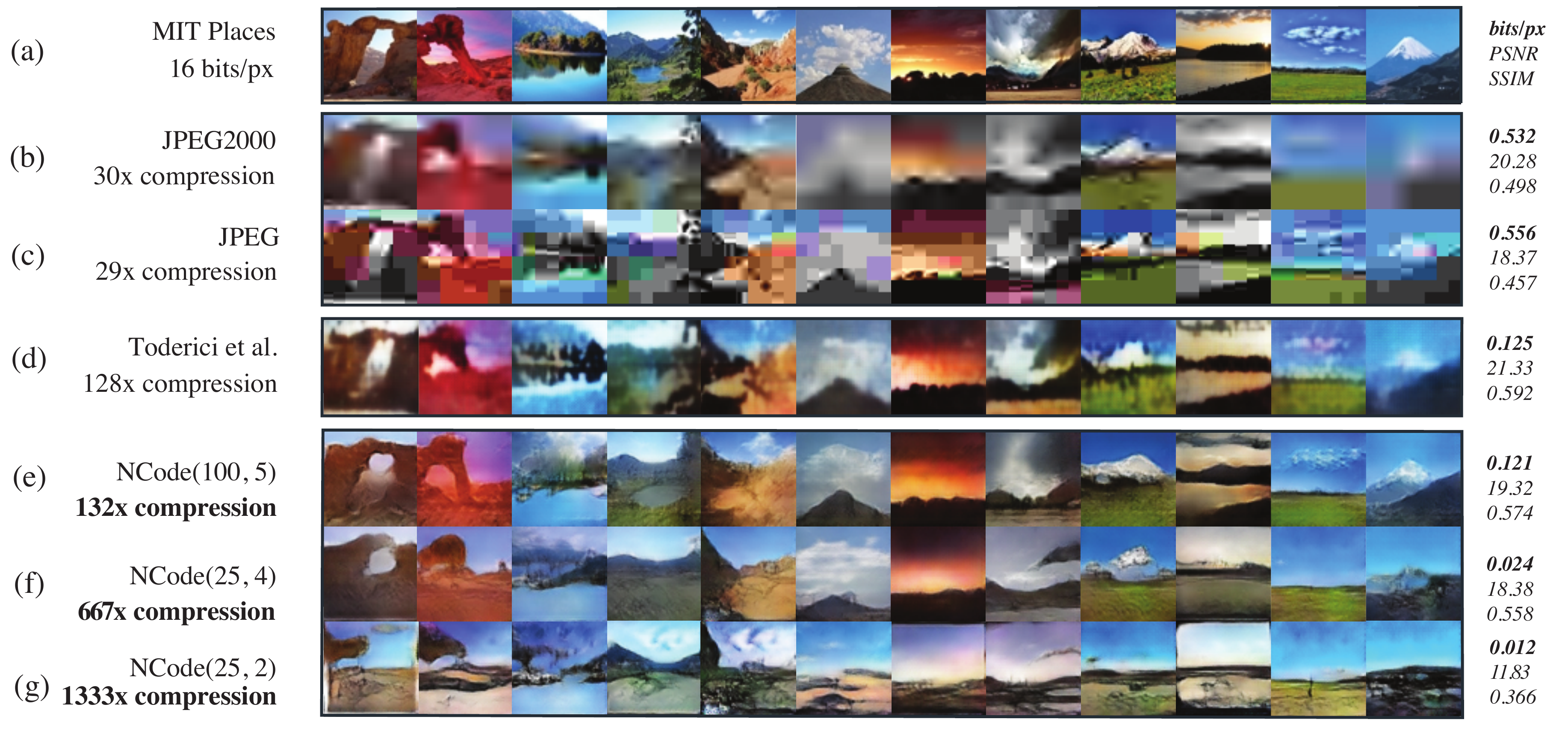} 
	\caption{A comparison of image reconstruction under various compression techniques. (a) Randomly sampled images from the  CelebA dataset (\textit{top panel})~\cite{liu2015faceattributes}, Zappos50k dataset (\textit{middle panel})~\cite{fine-grained} and Outdoor MIT Places dataset (\textit{bottom panel})~\cite{zhou2014learning}. Rows (b-d) demonstrate the corresponding reconstructions, compression ratios and PSNR/SSIM metrics (averaged over the full test set) for JPEG2000, JPEG and the thumbnail compression approach of Toderici et al. \cite{toderici2015variable,toderici2016full} respectively. Corresponding NCode performance is shown for varying latent vector dimension and quantization levels: (e) (100, 5 bit), (f) (25, 4 bit), and (g) (25, 2 bit) latent representations of the latent vector. File header size was excluded from JPEG/2000 results for fairer comparison. Best viewed zoomed in.}
	\label{fig:cifar}
\end{figure*}

We selected the CelebA~\cite{liu2015faceattributes}, UT Zappos50K~\cite{fine-grained} and MIT Places (outdoor natural scenes)~\cite{zhou2014learning} datasets for compression benchmarking. These datasets have the necessary data volume required for training, and are small enough ($64\times64$) to use with current GAN training (see Section 5). Traditional compression benchmarks, such as the Kodak PhotoCD dataset~ \cite{kodak}, currently fail on both criteria. Moreover, patch-based approaches to piecewise compression are unable to capture the image-level semantics that allow an image to be efficiently represented in terms of a low-dimensional latent vector from a generative model. We further evaluate our model on the popular CIFAR-10 dataset~\cite{Krizhevsky09learningmultiple}. Beyond data volume, the advantages of CIFAR are (a) that each example has a ground-truth class label (useful for validation), and (b) that it is one of very few large-scale image datasets to adopt lossless PNG compression. For MCode video compression, we select two categories (hand-waving and boxing) from the KTH actions dataset~\cite{schuldt2004recognizing}.

The encoder ($f_\theta(\mathbf{x})$), decoder ($g_\phi(\mathbf{z})$) and discriminator ($d_\vartheta(\mathbf{x})$) functions are all implemented as deep ConvNets~\cite{krizhevsky2012imagenet}. The decoder (generator) and discriminator networks adopt the standard DCGAN architecture of multiple convolutional layers with ReLU activation and batch normalization, which was shown to empirically improve training convergence~\cite{radford2015unsupervised}. The encoder network is identical to the discriminator, except for the output layer which produces a length-$M$ latent vector rather than a scalar in $[0,1]$. We vary the latent vector length $M=\{25, 50, 100 \}$ and sample from the uniform prior, $\mathcal{U}[-1,1]$. Substituting this for a truncated normal distribution had no notable impact.

Each NCode image dataset is partitioned into separate training and evaluation sets. For MCode video compression, we use the whole duration of $75\%$ and the first half of the remaining $25\%$ of videos for training, and the second half of that $25\%$ for evaluation. We use the Adam optimizer~\cite{kingma2014adam} with learning rate $0.0002$ and momentum $0.5$, and weight the pixel and perceptual loss terms with $\lambda_1 = 1$ and $\lambda_2 = 0.002$ respectively.

\subsection{NCode Image Compression}
We use NCode to compress and reconstruct images for each dataset and compare performance with JPEG/2000 (ImageMagick toolbox). As these schemes are not designed specifically for small images, we also compare to the state-of-the-art system for thumbnail compression presented by Toderici et al.~\cite{toderici2015variable,toderici2016full}. Performance is evaluated using the standard PSNR and SSIM metrics, averaged over the held-aside test set images. As these measures are known to correlate quite poorly with human perception of visual quality~\cite{toderici2016full, ledig2016photo}, we provide randomly-sampled images under each scheme in Figure~\ref{fig:cifar} to visually validate reconstruction performance. We also leverage the class labels associated with CIFAR-10 to propose an additional evaluation metric, i.e. the classification performance for a ConvNet independently trained on uncompressed examples. As file headers are responsible for a non-trivial portion of file size for small images, these were deducted when calculating compression for JPEG/JPEG2000. Huffman coding and quantization tables were however included for JPEG.


{Our results are presented in Figure~\ref{fig:cifar}. For each panel, row (a) presents raw image samples, (b-d) their JPEG2000, JPEG and Toderici et al. \cite{toderici2015variable, toderici2016full} reconstructions, and (e-g) their reconstructions using our proposed NCode method. To illustrate how NCode sample quality distorts with diminishing file size, we present sample reconstructions at varying latent vector length and quantization levels. These specific values were chosen to demonstrate (e) improved visual quality at similar compression levels, and (f-g) graceful degradation at extreme compression levels. It is clear that NCode(100,5) (length-100 latent representation, at 5 bits per vector entry) yields higher quality reconstructions (in terms of SSIM and visual inspection) than JPEG/2000 at $\sim 4$-fold higher compression levels. This compression ratio can be increased to a full order-of-magnitude greater than JPEG/2000 for NCode(25, 4) while still maintaining recognizable reconstructions. Even for the failure case of over-compression, NCode(25, 2) typically produces images that are plausible with respect to the underlying class semantics. NCode(100, 5) samples appear to be visually sharper and with fewer unnatural artifacts compared to the Toderici et al. approach set to maximum allowable compression.

Our appraisal of improved perceptual quality is supported by training a ConvNet to classify uncompressed CIFAR-10 images into their ten constituent categories, and observing how its accuracy drops when presented with images compressed under each scheme. Figure~\ref{tab:cifar} demonstrates that even using NCode(25, 4) ($\sim190$-fold compression), images are more recognizable than under the $\sim150$-fold compression of the Toderici el al. approach or $\sim15$-fold compression of JPEG/2000.
\begin{figure}[!t]
\begin{subfigure}[b]{0.45\textwidth}
	\begin{center}
		\renewcommand{\arraystretch}{1.2}
		\setlength\tabcolsep{4 pt}
		\begin{tabular}{|c|ccc|}
			\hline
			CIFAR-10 & \begin{tabular}[c]{@{}c@{}}bits/pixel\\ \end{tabular} & \begin{tabular}[c]{@{}c@{}}$\eta$\\ \end{tabular} & \begin{tabular}[c]{@{}c@{}} accuracy\\ \end{tabular} \\ \hline
			Original  & 19   & --                                                   & 70.53                                                         \\ \hline
			JPEG2000  &1.354   & 14                                                         & 37.48                                                        \\
			JPEG      & 1.962   & 10                                                        & 32.61                                                        \\ 
			Toderici et al.      & 0.125   & 152                                                         & 34.15                                                        \\ \hline
			NCode(100, 5) & 0.4883   & {39}                                                           & {54.96}                                                         \\
			NCode(25, 4)  & 0.0977   & {194}                                                           & {39.46}                                                         \\
			NCode(25, 2)  & 0.0488   & 389                                                           & 18.95    \\    \hline                                                
		\end{tabular}
		\vskip 0.1in
		\caption{}
		\label{tab:cifar}
	\end{center}
\end{subfigure}
\hfill
\begin{subfigure}[b]{0.475\textwidth}
	\centering
	\includegraphics[width=5.8cm]{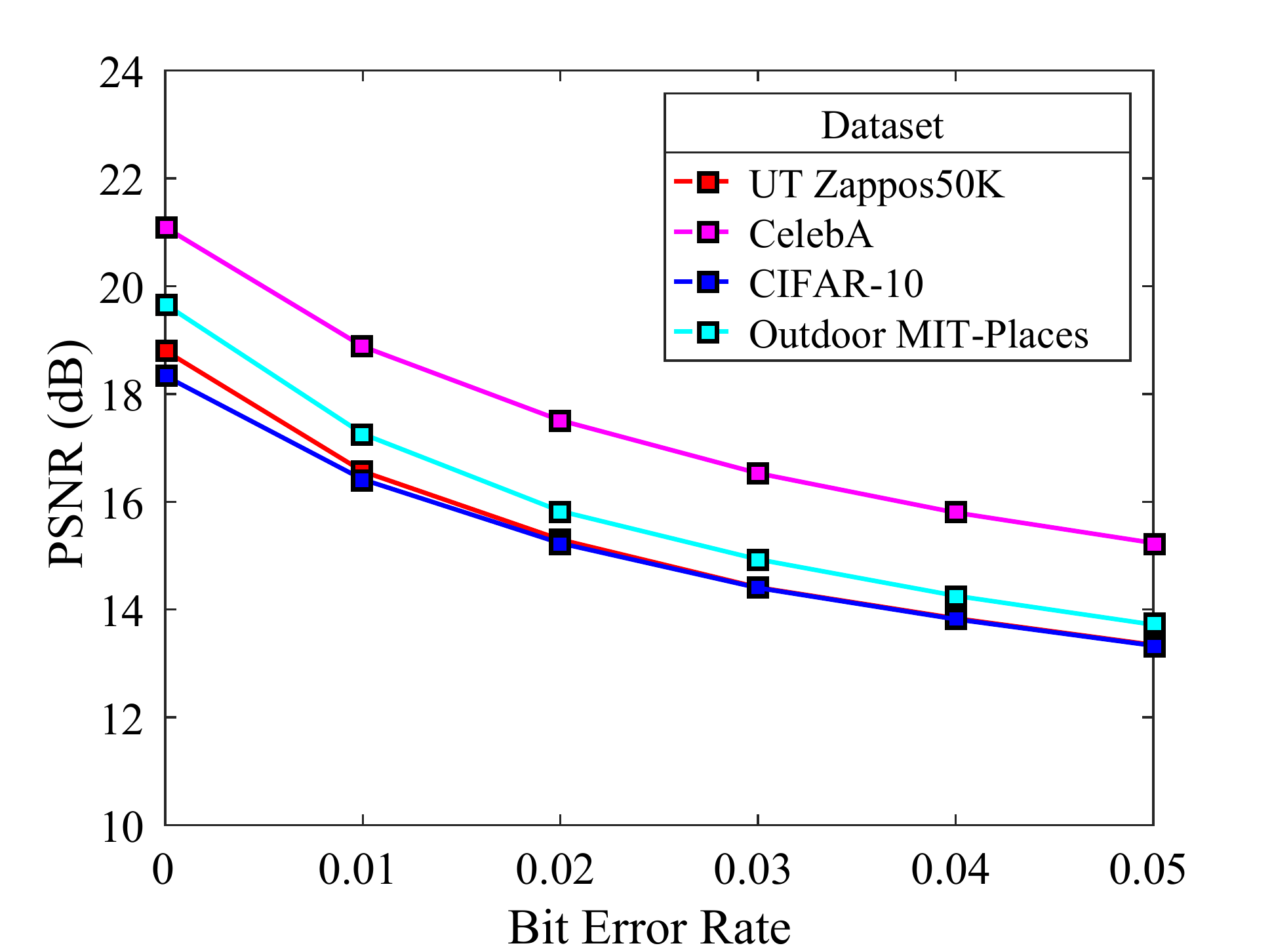}
	\caption{}
	\label{fig:ber}
\end{subfigure}
\caption{(a) Validation of perceptual plausibility of images compressed ($\eta$=compression factor) with NCode versus headerless JPEG, headerless JPEG2000 and Toderici et al \cite{toderici2015variable}, using a ConvNet independently trained on uncompressed images to categorize each sample into its respective CIFAR-10~\cite{Krizhevsky09learningmultiple} class. (b) Reduction in reconstruction quality (PSNR) as a function of bit error rate, $\varepsilon$, for each NCode image dataset. JPEG PSNR is known to degrade by more than $7$dB at $\varepsilon~\sim10^{-4}$.}
\vskip -0.1in
\end{figure}
\subsection{Robustness to Noisy Channels}
\label{ss:robustness}
The experiments presented above have assumed that $\mathbf{z}$ is transferred losslessly, with sender and receiver operating on a single machine. Where wireless signals are involved and in the absence of explicit error correction, bit error rates often occur with a frequency in the order of $\varepsilon = 10^{-3}$. It is also well established that traditional compression algorithms are not robust against these conditions, e.g. bit error rates in the order of just $\varepsilon = 10^{-4}$ result in unacceptable image distortion and a drop in PSNR of more than $7$dB~\cite{ho1997image,santa2000analytical,weerackody1996transmission}. 
The lack of robustness for traditional codecs is largely due to the introduction of variable-length entropy coding schemes, whereby the transmitted signal is essentially a map key with no preservation of semantic similarity between numerically adjacent signals. By contrast, the NCode system transmits explicit coordinates in the latent space $Z$ and thus should be robust against bit errors in $\mathbf{z}$, as shown in Figure~\ref{fig:ber}. Even at bit error rates of $\varepsilon = 10^{-2}$, which is greater than one should experience in practice, PSNR degrades by just $\sim 1$dB. 
\subsection{MCode Video Compression}
\label{ss:videos}
\begin{figure}[t!]
	\includegraphics[width=0.97\textwidth]{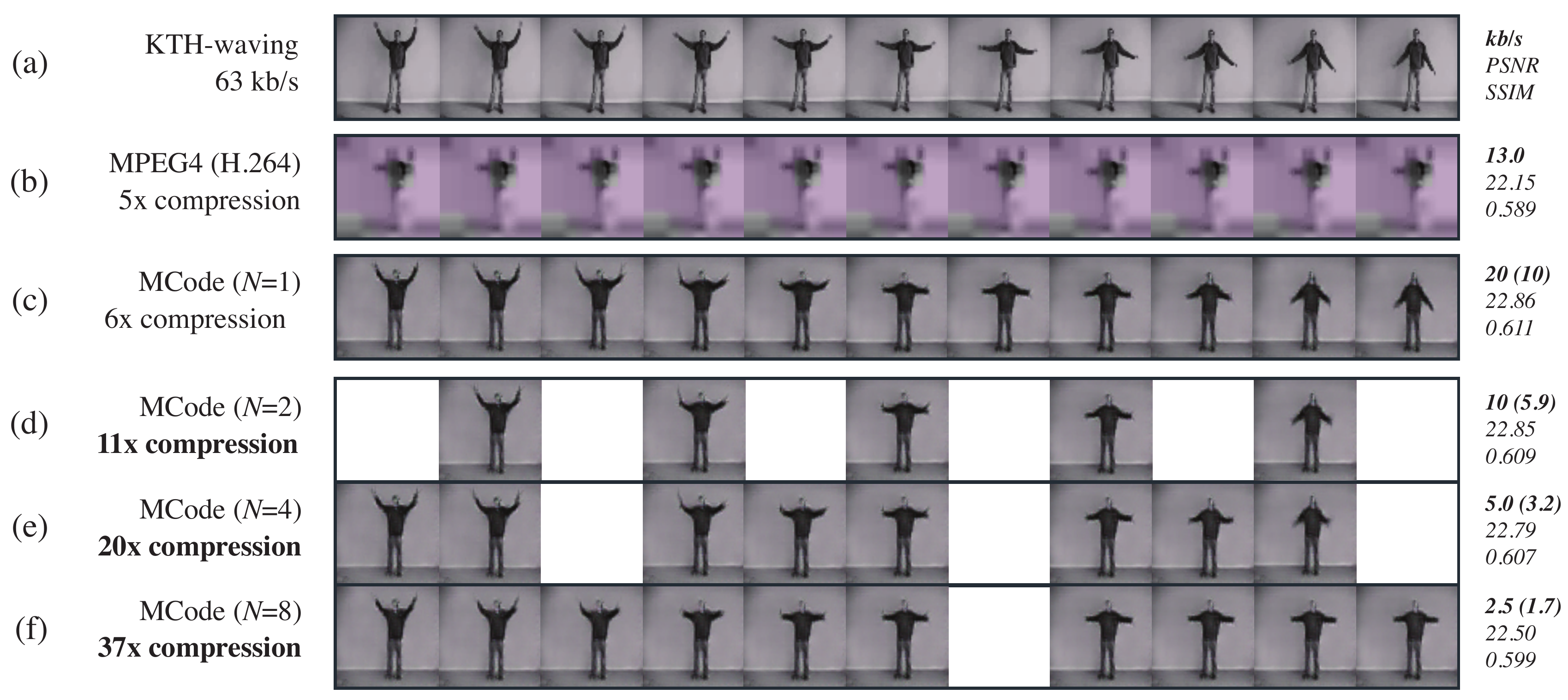}
	\caption{(a) Hand-waving video sequence randomly sampled from the KTH actions dataset~\cite{schuldt2004recognizing}. Row (b) demonstrates the corresponding frame-by-frame reconstructions, bitrates and mean PSNR/SSIM metrics (averaged over the full test set) for MPEG4 (H.264). Row (c) shows the corresponding performance for MCode using $N=1$, i.e. applying image NCode frame-by-frame. Rows (d-f) demonstrate the extra performance than can be leveraged by linear interpolation between latent vectors $\mathbf{z}^{(t)}$ and $\mathbf{z}^{(t+N)}$ for (d) $N=2$, (e) $N=4$ and (f) $N=8$ (transmitted frames omitted). Bit rates are presented both before and after Huffman coding (parentheses). Best viewed zoomed in.}
	\label{fig:boxing}
	\vskip -0.15in
\end{figure}
{We apply MCode to compress and reconstruct frames from the aforementioned KTH dataset and compare performance against the MPEG4 (H.264) codec (FFMPEG toolbox). Similar to image compression, performance is evaluated against mean frame-wise PSNR and SSIM metrics (average over test videos) and visualizations provided in Figure~\ref{fig:boxing} for the handwaving dataset. Results from the boxing dataset are similar and included in the Supplementary Material. Comparing (b) MPEG to (c) frame-by-frame MCode, it is clear that our method provides higher quality results at a comparable compression level.  Despite similar PSNR, the relative preservation of background texture and limb sharpness is noteworthy.}

{Motivated by MPEG bidirectional prediction, MCode can produce greater compression by interpolating between frames in latent space. This process is shown in Figure~\ref{fig:boxing} (d-f). Frames transmitted and reconstructed using standard NCode are are omitted, with the remaining $N-1$ interpolated frames shown for (d) $N = 2$, (e) $N = 4$ and (f) $N = 8$. These temporal correlations can be further leveraged by transmitting the Huffman-encoded difference between $\mathbf{z}^{(t)}$ and $\mathbf{z}^{(t+N)}$, leading to a further $20\%$-$50\%$ lossless compression on average. As shown in Figure~\ref{fig:boxing}, this can lead to order-of-magnitude reduction in bitrate over MPEG4 while providing more visually plausible sequences.}

The robustness analysis in Section~\ref{ss:robustness} extends to video MCode in the absence of inter-frame entropy coding. Figure~\ref{fig:boxing} presented MCode compression factors both with and without Huffman coding, which introduced $20\%$-$50\%$ further performance improvement on average. We present both options for the user to choose based on the robustness versus compression constraints of their application.
\section{Large Image Compression}
\label{sec:largeimages}

\remove{
In this paper we have demonstrated a proof-of-concept that adversarially trained neural codecs can yield an order-of-magnitude improvement in image and video compression when compared to traditional compression schemes, and that as compression levels increase, the transmitted data degrades in a graceful manner. Although this is a promising result, it raises the obvious question of whether these results can generalize to less trivial examples. Hand-crafted codecs such as JPEG2000 and MPEG4 are not restricted by image size or class semantics, but our NCode and MCode models were so far only demonstrated for relatively simple $32$x$32$ and $64$x$64$ examples.
}

So far we have only demonstrated generative compression for relatively simple $64$x$64$ examples. Although our results are promising, this raises the obvious question of whether these results can generalize to larger images with more complex class sematics, e.g. those represented by ImageNet. In fact, we believe that the compression factors presented here should only continue to improve for larger images. The latent vector $\textbf{z}$ describes semantic content of an image, which should grow sub-linearly (or likely remain constant) for higher-resolution images of equivalent content. 

Current limitations in this direction are not fundamental to generative compression, but are instead those of generative modelling more generally. Although autoencoders can certainly be extended to larger images, as demonstrated by the beautiful reconstructions of~\cite{Theis2017a}, GANs fail for large images due to well-established instabilities in the adversarial training process~\cite{theis2015note}. Thankfully this is also an area receiving substantial attention, and in Figure~\ref{fig:wganfig} we demonstrate the impact of the past year of GAN research on 256x256 image compression (held-aside test samples from a 10-class animal subset of ImageNet). Unlike vanilla DCGAN (used throughout this paper), the more recent Wasserstein GAN~\cite{arjovsky2017wasserstein} produces recognizable reconstructions. Likewise, this and other improved generative models could be leveraged to improve the compression factors and reconstruction quality demonstrated throughout this paper. We expect this field will continue to improve at a rapid pace.
\begin{figure}[t!]
	\begin{center}
		\centerline{\includegraphics[width=1.0\columnwidth]{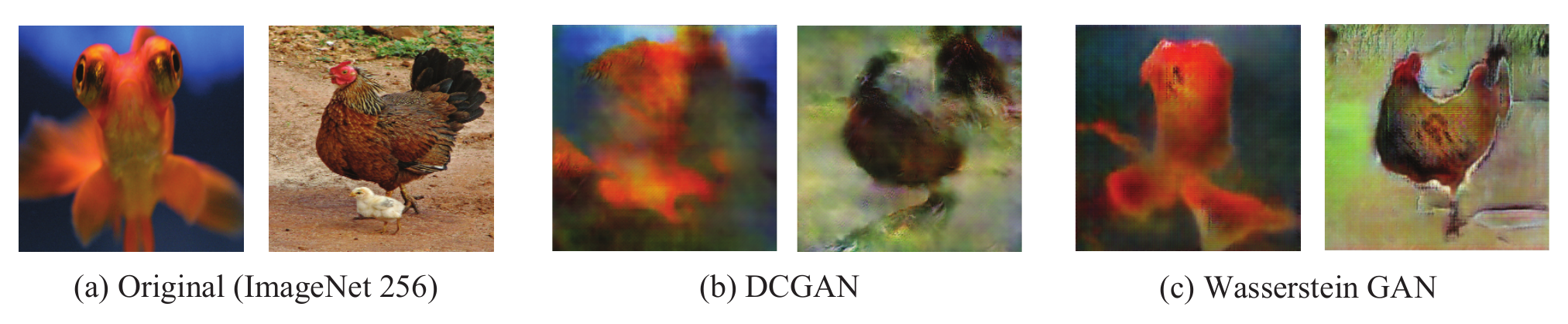}}
		\caption{The impact of one year of GAN advancements -- (b) DCGAN~\cite{radford2015unsupervised} to (c) Wasserstein GAN~\cite{arjovsky2017wasserstein} -- on generative compression performance (256x256 ImageNet). Best viewed zoomed in.}
		\label{fig:wganfig}
	\end{center}
	\vskip -0.2in
\end{figure} 

\section{Discussion}
\label{sec:discussion}
\remove{We adopted a simple DCGAN for this purpose, though any suitable generative model conditioned on a latent representation could be applied.}
All compression algorithms involve a pair of analysis and synthesis transforms that aim to accurately reproduce the original images. Traditional, hand-crafted codecs lack adaptability and are unable to leverage semantic redundancy in natural images. Moreover, earlier neural network-based approaches optimize against a pixel-level objective that tends to produce blurry reconstructions. In this paper we propose generative compression as an alternative, where we first train the synthesis transform as a generative model. We adopted a simple DCGAN for this purpose, though any suitable generative model conditioned on a latent representation could be applied. This synthesis transform is then used as a non-adaptive decoder in an autoencoder setup, thereby confining the search space of reconstructions to a smaller compact set of natural images enriched for the appropriate class semantics. 
\begin{figure}[h!]
	\includegraphics[width=0.97\textwidth]{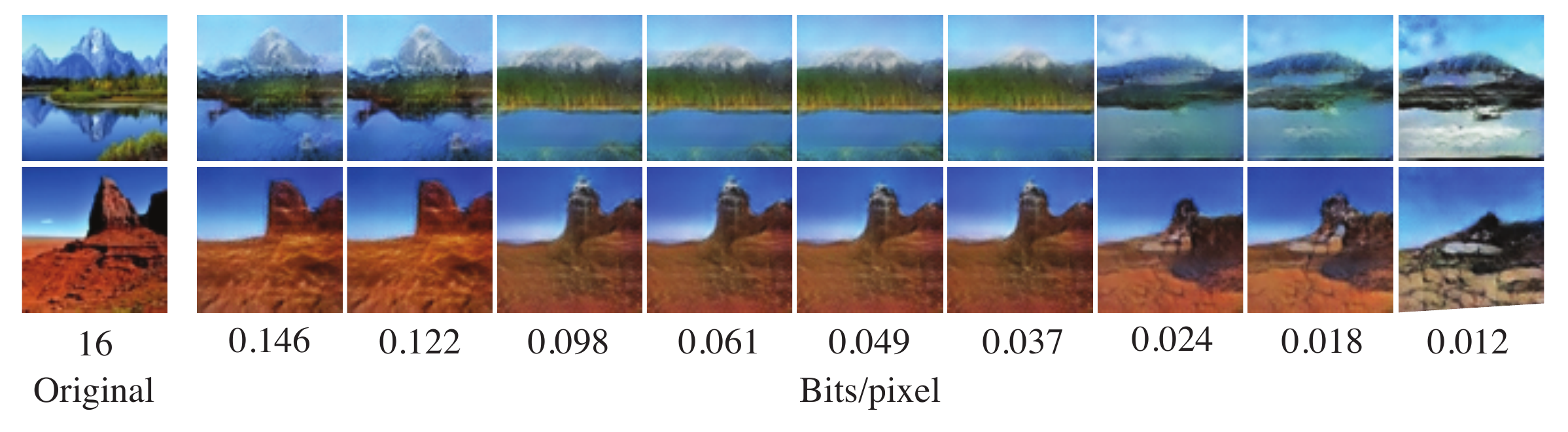}
	\caption{Graceful degradation of generative compression at 2-to-3 orders-of-magnitude compression.}
	\label{fig:gd}
	\vskip -0.1in
\end{figure}

We have demonstrated the {potential} of generative compression for orders-of-magnitude improvement in image and video compression -- both in terms of compression factor and noise tolerance -- when compared to traditional schemes. Generatively compressed images also degrade gracefully at deeper compression levels, as shown in Figure~\ref{fig:gd}. These results are possible because the transmitted data is merely a description with respect to the receiver's shared understanding of natural image semantics.

To implement generative compression as a practical compression standard, we envision that devices could maintain a synchronized collection of cached manifolds that capture the low-dimensional embeddings of common objects and scenes (e.g. the 1000 ImageNet categories). In much the same way that the JPEG codec is not transmitted alongside every JPEG-compressed image, this removes the need to explicitly transmit network weights for previously encountered concepts. This collection could be further augmented by and individual's usage patterns, e.g. to learn rich manifolds over the appearance of rooms and faces that regularly appear via online video calls. We believe that such a system would better replicate the efficiency with which humans can store and communicate memories of complex scenes, and would be a valuable component of any generally intelligent system.

\clearpage
\section*{Acknowledgments}
Support is gratefully acknowledged from the National Science Foundation (NSF) under grants IIS-1447786 and CCF-1563880, and the Intelligence Advanced Research Projects Activity (IARPA) under grant 138076-5093555.
\bibliography{refs}

\begin{thebibliography}{10}

\bibitem{gregor2016towards}
Karol Gregor, Frederic Besse, Danilo~Jimenez Rezende, Ivo Danihelka, and Daan
  Wierstra.
\newblock Towards conceptual compression.
\newblock In {\em Advances In Neural Information Processing Systems}, pages
  3549--3557, 2016.

\bibitem{goodfellow2014generative}
Ian Goodfellow, Jean Pouget-Abadie, Mehdi Mirza, Bing Xu, David Warde-Farley,
  Sherjil Ozair, Aaron Courville, and Yoshua Bengio.
\newblock Generative adversarial nets.
\newblock In {\em Advances in neural information processing systems}, pages
  2672--2680, 2014.

\bibitem{gersho1992vector}
Allen Gersho and Robert~M Gray.
\newblock Vector quantization i: Structure and performance.
\newblock In {\em Vector quantization and signal compression}, pages 309--343.
  Springer, 1992.

\bibitem{Theis2017a}
L.~Theis, W.~Shi, A.~Cunningham, and F.~Huszar.
\newblock Lossy image compression with compressive autoencoders.
\newblock In {\em International Conference on Learning Representations}, 2017.

\bibitem{balle2016end}
Johannes Ball{\'e}, Valero Laparra, and Eero~P Simoncelli.
\newblock End-to-end optimized image compression.
\newblock {\em arXiv preprint arXiv:1611.01704}, 2016.

\bibitem{toderici2015variable}
George Toderici, Sean~M O'Malley, Sung~Jin Hwang, Damien Vincent, David Minnen,
  Shumeet Baluja, Michele Covell, and Rahul Sukthankar.
\newblock Variable rate image compression with recurrent neural networks.
\newblock {\em arXiv preprint arXiv:1511.06085}, 2015.

\bibitem{toderici2016full}
George Toderici, Damien Vincent, Nick Johnston, Sung~Jin Hwang, David Minnen,
  Joel Shor, and Michele Covell.
\newblock Full resolution image compression with recurrent neural networks.
\newblock {\em arXiv preprint arXiv:1608.05148}, 2016.

\bibitem{oord2016pixel}
Aaron van~den Oord, Nal Kalchbrenner, and Koray Kavukcuoglu.
\newblock Pixel recurrent neural networks.
\newblock {\em arXiv preprint arXiv:1601.06759}, 2016.

\bibitem{theis2015generative}
Lucas Theis and Matthias Bethge.
\newblock {Generative image modeling using spatial LSTMs}.
\newblock In {\em Advances in Neural Information Processing Systems}, pages
  1927--1935, 2015.

\bibitem{kingma2013auto}
Diederik~P Kingma and Max Welling.
\newblock Auto-encoding variational bayes.
\newblock {\em arXiv preprint arXiv:1312.6114}, 2013.

\bibitem{larsen2015autoencoding}
Anders Boesen~Lindbo Larsen, S{\o}ren~Kaae S{\o}nderby, Hugo Larochelle, and
  Ole Winther.
\newblock Autoencoding beyond pixels using a learned similarity metric.
\newblock {\em arXiv preprint arXiv:1512.09300}, 2015.

\bibitem{radford2015unsupervised}
Alec Radford, Luke Metz, and Soumith Chintala.
\newblock Unsupervised representation learning with deep convolutional
  generative adversarial networks.
\newblock {\em arXiv preprint arXiv:1511.06434}, 2015.

\bibitem{gupta2015deep}
Suyog Gupta, Ankur Agrawal, Kailash Gopalakrishnan, and Pritish Narayanan.
\newblock Deep learning with limited numerical precision.
\newblock In {\em ICML}, pages 1737--1746, 2015.

\bibitem{hubara2016quantized}
Itay Hubara, Matthieu Courbariaux, Daniel Soudry, Ran El-Yaniv, and Yoshua
  Bengio.
\newblock Quantized neural networks: Training neural networks with low
  precision weights and activations.
\newblock {\em arXiv preprint arXiv:1609.07061}, 2016.

\bibitem{ledig2016photo}
Christian Ledig, Lucas Theis, Ferenc Huszar, Jose Caballero, Andrew Cunningham,
  Alejandro Acosta, Andrew Aitken, Alykhan Tejani, Johannes Totz, Zehan Wang,
  et~al.
\newblock Photo-realistic single image super-resolution using a generative
  adversarial network.
\newblock {\em arXiv preprint arXiv:1609.04802}, 2016.

\bibitem{gatys2015texture}
Leon Gatys, Alexander~S Ecker, and Matthias Bethge.
\newblock Texture synthesis using convolutional neural networks.
\newblock In {\em Advances in Neural Information Processing Systems}, pages
  262--270, 2015.

\bibitem{dosovitskiy2016generating}
Alexey Dosovitskiy and Thomas Brox.
\newblock Generating images with perceptual similarity metrics based on deep
  networks.
\newblock In {\em Advances in Neural Information Processing Systems}, pages
  658--666, 2016.

\bibitem{gatys2016image}
Leon~A Gatys, Alexander~S Ecker, and Matthias Bethge.
\newblock Image style transfer using convolutional neural networks.
\newblock In {\em Proceedings of the IEEE Conference on Computer Vision and
  Pattern Recognition}, pages 2414--2423, 2016.

\bibitem{krizhevsky2012imagenet}
Alex Krizhevsky, Ilya Sutskever, and Geoffrey~E Hinton.
\newblock Imagenet classification with deep convolutional neural networks.
\newblock In {\em Advances in neural information processing systems}, pages
  1097--1105, 2012.

\bibitem{zhu2016generative}
Jun-Yan Zhu, Philipp Kr{\"a}henb{\"u}hl, Eli Shechtman, and Alexei~A Efros.
\newblock Generative visual manipulation on the natural image manifold.
\newblock In {\em European Conference on Computer Vision}, pages 597--613.
  Springer, 2016.

\bibitem{lamb2016discriminative}
Alex Lamb, Vincent Dumoulin, and Aaron Courville.
\newblock Discriminative regularization for generative models.
\newblock {\em arXiv preprint arXiv:1602.03220}, 2016.

\bibitem{donahue2016adversarial}
Jeff Donahue, Philipp Kr{\"a}henb{\"u}hl, and Trevor Darrell.
\newblock Adversarial feature learning.
\newblock {\em arXiv preprint arXiv:1605.09782}, 2016.

\bibitem{dumoulin2016adversarially}
Vincent Dumoulin, Ishmael Belghazi, Ben Poole, Alex Lamb, Martin Arjovsky,
  Olivier Mastropietro, and Aaron Courville.
\newblock Adversarially learned inference.
\newblock {\em arXiv preprint arXiv:1606.00704}, 2016.

\bibitem{budden2016deep}
David Budden, Alexander Matveev, Shibani Santurkar, Shraman~Ray Chaudhuri, and
  Nir Shavit.
\newblock Deep tensor convolution on multicores.
\newblock {\em arXiv preprint arXiv:1611.06565}, 2016.

\bibitem{shi2016real}
Wenzhe Shi, Jose Caballero, Ferenc Huszar, Johannes Totz, Andrew~P Aitken, Rob
  Bishop, Daniel Rueckert, and Zehan Wang.
\newblock Real-time single image and video super-resolution using an efficient
  sub-pixel convolutional neural network.
\newblock In {\em Proceedings of the IEEE Conference on Computer Vision and
  Pattern Recognition}, pages 1874--1883, 2016.

\bibitem{brock2016neural}
Andrew Brock, Theodore Lim, JM~Ritchie, and Nick Weston.
\newblock Neural photo editing with introspective adversarial networks.
\newblock {\em arXiv preprint arXiv:1609.07093}, 2016.

\bibitem{liu2015faceattributes}
Ziwei Liu, Ping Luo, Xiaogang Wang, and Xiaoou Tang.
\newblock Deep learning face attributes in the wild.
\newblock In {\em Proceedings of International Conference on Computer Vision
  (ICCV)}, 2015.

\bibitem{fine-grained}
A.~Yu and K.~Grauman.
\newblock {F}ine-{G}rained {V}isual {C}omparisons with {L}ocal {L}earning.
\newblock In {\em Computer Vision and Pattern Recognition (CVPR)}, June 2014.

\bibitem{zhou2014learning}
Bolei Zhou, Agata Lapedriza, Jianxiong Xiao, Antonio Torralba, and Aude Oliva.
\newblock Learning deep features for scene recognition using places database.
\newblock In {\em Advances in neural information processing systems}, pages
  487--495, 2014.

\bibitem{kodak}
Kodak.
\newblock Kodak lossless true color image suite.
\newblock \url{http://r0k.us/graphics/kodak/}, 1999.

\bibitem{Krizhevsky09learningmultiple}
Alex Krizhevsky.
\newblock Learning multiple layers of features from tiny images.
\newblock Technical report, 2009.

\bibitem{schuldt2004recognizing}
Christian Schuldt, Ivan Laptev, and Barbara Caputo.
\newblock Recognizing human actions: A local svm approach.
\newblock In {\em Pattern Recognition, 2004. ICPR 2004. Proceedings of the 17th
  International Conference on}, volume~3, pages 32--36. IEEE, 2004.

\bibitem{kingma2014adam}
Diederik Kingma and Jimmy Ba.
\newblock Adam: A method for stochastic optimization.
\newblock {\em arXiv preprint arXiv:1412.6980}, 2014.

\bibitem{ho1997image}
Keang-Po Ho and Joseph~M Kahn.
\newblock Image transmission over noisy channels using multicarrier modulation.
\newblock {\em Signal Processing: Image Communication}, 9(2):159--169, 1997.

\bibitem{santa2000analytical}
Diego Santa-Cruz and Touradj Ebrahimi.
\newblock An analytical study of jpeg 2000 functionalities.
\newblock In {\em Image Processing, 2000. Proceedings. 2000 International
  Conference on}, volume~2, pages 49--52. IEEE, 2000.

\bibitem{weerackody1996transmission}
Vijitha Weerackody, Christine Podilchuk, and Anthony Estrella.
\newblock Transmission of jpeg-coded images over wireless channels.
\newblock {\em Bell Labs Technical Journal}, 1(2):111--126, 1996.

\bibitem{theis2015note}
Lucas Theis, A{\"a}ron van~den Oord, and Matthias Bethge.
\newblock A note on the evaluation of generative models.
\newblock {\em arXiv preprint arXiv:1511.01844}, 2015.

\bibitem{arjovsky2017wasserstein}
Martin Arjovsky, Soumith Chintala, and L{\'e}on Bottou.
\newblock {Wasserstein GAN}.
\newblock {\em arXiv preprint arXiv:1701.07875}, 2017.

\end{thebibliography}
\bibliographystyle{unsrt}
\end{document}